\documentclass{edm_article}

\usepackage{booktabs} 
\usepackage{xcolor}

\begin{document}

\title{From Flat to Structural: Enhancing Automated Short Answer Grading with GraphRAG}


%
%
%
%

\numberofauthors{9} 
\author{
\alignauthor Yucheng Chu \\
       \affaddr{Michigan State University}\\
       \email{chuyuch2@msu.edu}
\alignauthor Haoyu Han \\
       \affaddr{Michigan State University}\\
       \email{hanhaoy1@msu.edu}
\alignauthor Shen Dong\\
        \affaddr{Michigan State University}\\
       \email{dongshe1@msu.edu	}
\and  
\alignauthor
Hang Li\\
       \affaddr{Michigan State University}\\
       \email{lihang4@msu.edu}
\alignauthor Kaiqi Yang\\
        \affaddr{Michigan State University}\\
       \email{kqyang@msu.edu}
\alignauthor Yasemin Copur-Gencturk\\
       \affaddr{University of Southern California}\\
       \email{copurgen@usc.edu}
\and
\alignauthor Joseph Krajcik\\
       \affaddr{Michigan State University}\\
       \email{krajcik@msu.edu}
\alignauthor Namsoo Shin\\
       \affaddr{Michigan State University}\\
       \email{namsoo@msu.edu}
\and 
\alignauthor Hui Liu\\
       \affaddr{Michigan State University}\\
       \email{liuhui7@msu.edu}
}

\maketitle

\begin{abstract}


Automated short answer grading (ASAG) is critical for scaling educational assessment, yet large language models (LLMs) often struggle with hallucinations and strict rubric adherence due to a mismatch between generalized pre-training objectives and the task-specific requirements of grading. While Retrieval-Augmented Generation (RAG) can reduce hallucinations by providing access to relevant reference materials, standard flat vector retrieval treats knowledge as isolated fragments, failing to capture the structural relationships and multi-hop dependencies inherent in complex educational content. To explore whether incorporating structural information is beneficial for ASAG, we investigate Graph Retrieval-Augmented Generation (GraphRAG) as a general paradigm and systematically evaluate two representative instantiations, Microsoft GraphRAG and HippoRAG, on this task. These approaches differ in how they construct and traverse knowledge graphs, providing complementary perspectives on graph-based evidence retrieval. Experimental evaluations on a Next Generation Science Standards (NGSS) dataset demonstrate that structural retrieval significantly outperforms standard RAG baselines across all metrics. Notably, the HippoRAG achieved substantial improvements in evaluating Science and Engineering Practices (SEP), confirming the superiority of structural retrieval in verifying the logical reasoning chains required for higher-order academic assessment.
\end{abstract}


\section{Introduction}

Automated short answer grading (ASAG) has emerged as a critical technology for scaling educational assessment, aiming to evaluate student responses with the same accuracy and consistency as human instructors. While early approaches relied on keyword matching and latent semantic analysis, the advent of large language models (LLMs) has fundamentally transformed the landscape. LLMs possess remarkable semantic understanding and generation capabilities. These features allow them to not only interpret complex student answers but also, crucially, to understand and adhere to nuanced grading rubrics with human-level flexibility. However, despite their fluency, off-the-shelf LLMs frequently struggle with the strict constraints of academic grading. They are prone to ``hallucinations'' where plausible but incorrect justifications are generated. Furthermore, they often fail to adhere strictly to specific grading rubrics, relying instead on their generalized internal knowledge acquired during pre-training. This vast but generic training data often conflicts with sepcific course definitions or omits the precise constraints required by a local curriculum. To mitigate these limitations, recent works have integrated retrieval-augmented generation (RAG) into the grading pipeline. A prominent example of this advancement is the system proposed by Chu et al., which enhances LLM grading by grounding the model's decisions in a specific knowledge base \cite{chu2025enhancing}. Their approach constructs a repository of reference answers and scoring criteria, utilizing a dual retrieval strategy to fetch the most relevant grading standards for a given student response. By dynamically providing the LLM with the exact rubric criteria needed for a specific question, this RAG-based method significantly reduces hallucination and improves the alignment between automated scores and human judgment. The system effectively converts the grading task from an open-ended generation problem into a grounded reasoning task, where the model must support its evaluation with retrieved evidence.

Despite the improvements offered by standard RAG frameworks, significant technical limitations remain when applied to complex educational content. Standard RAG systems typically rely on vector similarity search, which encodes text into high-dimensional semantic embeddings and retrieves the ``chunks" of information semantically related to the query via the mathematical distance calculations. This ``flat" retrieval method is popular adopted due to its simplicity and high efficiency. However, it fails to leverage the structural relationships between concepts because it treats knowledge as isolated fragments. To illustrate this limitation in ASAG, consider a grading scenario requiring a multi-hop reasoning chain: increased temperature (A) leads to enzyme denaturation (B), which in turn causes a cessation of metabolic reactions (C). If a student argues that ``high heat stops the reaction'' (linking $A$ directly to $C$), a standard RAG system will likely retrieve chunks containing ``temperature'' (A) and ``reaction rates'' (C) due to keyword similarity. However, it frequently fails to retrieve the distinct text describing ``denaturation'' ($B$) because the student's response lacks the specific terminology to trigger a match for that intermediate concept. Lacking this intermediate link, the retrieval context fails to bridge the causal gap between the stimulus and the result.
This fragmentation directly degrades the performance of automated grading. When the retrieval system fails to provide the bridging evidence ($B$), the LLM lacks the necessary structural context to verify the student's logic. Consequently, the model is forced to rely on its pre-trained internal parameters to infer the missing connection, reintroducing the risk of hallucinations or causing the model to incorrectly penalize a valid conceptual leap.

To address the inability of flat retrieval systems to capture deep semantic structure, this work introduces a novel grading framework based on graph retrieval-augmented generation (GraphRAG) \cite{han2025reasoning}. Unlike standard RAG, which indexes information as independent vectors, GraphRAG organizes the reference material and grading rubrics into a structured \textit{knowledge graph} (KG). In this architecture, specific concepts, criteria, and keywords act as nodes, while the logical relationships between them, such as hierarchy, causality, and prerequisites, are represented as edges. This graph-based structure allows the system to move beyond simple similarity matching. When evaluating a student's answer, our system does not merely fetch isolated text chunks; it traverses the knowledge graph to retrieve a connected subgraph of information. This enables the retrieval of ``multi-hop" evidence, where the system can identify that concept A is relevant to the student's answer because it is causally linked to concept B, even if concept A was not explicitly mentioned in the initial query. By preserving the relational context of the reference material, our approach provides the LLM with a holistic view of the grading criteria, rather than fragmented snapshots. Our method is particularly advantageous for questions requiring higher-order thinking, where the correctness of an answer depends on the synthesis of multiple concepts. The following sections detail the construction of the grading knowledge graph, the graph traversal algorithms employed during the grading phase, and the comparative experimental results. Our evaluations on Next Generation Science Standards (NGSS) tasks demonstrate that this structural retrieval approach significantly outperforms standard RAG baselines, with the HippoRAG architecture achieving particularly substantial improvements in verifying the logical reasoning chains required for Science and Engineering Practices.

\section{Related Work}

The development of automated short answer grading (ASAG) has evolved from simple lexical matching to complex semantic reasoning systems. This section reviews the historical progression of grading technologies, the recent integration of retrieval-augmented generation (RAG) to address hallucination, and the emerging paradigm of graph-based retrieval which forms the technical foundation of this work.

\subsection{Automated Short Answer Grading}
Early approaches to ASAG relied heavily on statistical and keyword-based methods. Techniques such as latent semantic analysis (LSA) and unigram overlap were employed to measure the similarity between a student's response and a reference answer \cite{foltz1999automatically, mohler2011learning}. While computationally efficient, these methods operated on a ``bag-of-words'' assumption, ignoring word order and syntactic structure, which often led to the misclassification of answers that used correct terminology in incorrect contexts. The field advanced significantly with the introduction of pre-trained language models (PLMs) such as BERT. These models utilized self-attention mechanisms to capture contextual nuances, allowing for more accurate semantic matching. However, BERT-based systems were primarily discriminative, treating grading as a classification task (correct vs. incorrect) rather than generating interpretable feedback. The advent of large language models (LLMs) shifted the paradigm towards generative grading, where models could both score an answer and provide a justification. Despite this capability, generative LLMs suffer from a lack of domain specificity and a tendency to hallucinate, fabricating plausible but factually incorrect reasoning when the model's internal knowledge conflicts with specific course rubrics.

\subsection{RAG in Education}
To mitigate the hallucination and context limitations of off-the-shelf LLMs, researchers began adopting RAG \cite{fan2024survey}. RAG systems decouple the model's reasoning capabilities from its knowledge storage by retrieving relevant documents from an external corpus and injecting them into the model's context window. This approach is particularly vital in educational settings where grading criteria are rigid and course-specific. The work of Chu et al. represents the state-of-the-art in this domain \cite{chu2025enhancing}. Their system utilizes a dense vector retrieval mechanism to fetch specific reference answers and scoring rubrics relevant to a student's query. By grounding the LLM's generation in these retrieved artifacts, they successfully demonstrated a reduction in grading errors and improved alignment with human evaluators. However, standard RAG systems, including the one proposed by Chu et al., rely on ``flat" retrieval strategies. They chunk documents into independent text segments and retrieve them based on vector similarity. This process treats knowledge as isolated fragments, often failing to retrieve information that is conceptually related but not explicitly similar in vector space, such as multi-step causal chains or hierarchical dependencies essential for grading complex questions.

\subsection{GraphRAG}
Graph retrieval-augmented generation (GraphRAG) has emerged as a solution to the limitations of flat retrieval by introducing structural awareness into the generation pipeline \cite{han2024retrieval}. Unlike standard RAG, which views data as an unstructured collection of text chunks, GraphRAG organizes information into a knowledge graph (KG), where entities (concepts) are nodes and relationships (dependencies, causality) are edges. Microsoft Research's recent implementation of GraphRAG demonstrates that traversing these graph structures allows LLMs to perform ``global" reasoning over a dataset, connecting disparate pieces of information that standard vector search would miss \cite{edge2024from}. In the context of grading, this allows a system to recognize that a student's answer regarding a specific biological function is correct not just because it matches a keyword, but because it correctly identifies the prerequisite process defined in the knowledge graph. By traversing the edges of the graph, the retrieval module can fetch a subgraph containing the question, the direct answer, and the underlying principles, providing the LLM with a structurally complete context for evaluation.

\section{Problem Statement}

While the integration of retrieval-augmented generation (RAG) into grading pipelines has significantly reduced the incidence of hallucination, current state-of-the-art systems remain constrained by their underlying retrieval mechanisms. The fundamental limitation lies in the mismatch between the ``flat" data representation used by standard vector retrievers and the highly structured, relational nature of educational knowledge.

\subsection{Limitations of Flat Vector Retrieval}
Standard RAG systems, including the baseline framework proposed by Chu et al. \cite{chu2025enhancing}, operate on the principle of semantic similarity matching. In this paradigm, the grading rubric and reference materials are segmented into discrete text chunks, $D = \{d_1, d_2, ..., d_n\}$, which are then encoded into high-dimensional vector embeddings. Given a student's answer as a query $q$, the system retrieves a subset of chunks based on a similarity metric, such as cosine similarity, expressed as $Score(q, d_i) = \cos(\mathbf{v}_q, \mathbf{v}_{d_i})$.

This approach relies on the assumption that the necessary grading criteria are semantically proximate to the student's answer in vector space. However, this assumption frequently fails in educational contexts where concepts are linked by logical dependencies rather than mere lexical overlap. A student's response might correctly discuss a downstream effect (concept C) of a biological process (concept A) without explicitly mentioning the intermediate mechanism (concept B). A flat retriever, searching for terms related to A and C, is likely to retrieve chunks containing only A or C. It often fails to retrieve the bridging chunk containing concept B because B is not semantically similar enough to a query embedding constructed from A and C. Consequently, the LLM is presented with fragmented evidence so that it knows the cause and the effect but misses the crucial explanatory link required to verify the student's reasoning.

\subsection{The Challenge of Multi-Hop Reasoning}
The core deficiency of flat retrieval in automated short answer grading is the inability to support multi-hop reasoning. Academic grading often requires verifying a chain of logic: verifying that a student understands that $X \rightarrow Y \rightarrow Z$. In a standard vector store, the relationship $X \rightarrow Y$ and $Y \rightarrow Z$ may reside in separate, non-adjacent chunks. 
To quantify the necessity of this reasoning, we conducted a preliminary analysis of the NGSS grading dataset. Our examination revealed that approximately 40\% of the questions in the Science and Engineering Practices (SEP) dimension require the synthesis of information across at least two distinct conceptual nodes that are not co-located in the source text. For example, in Task 2 (Chemical Reactions in Soap Making), standard RAG frequently failed to grade responses correctly regarding the distinction between \textit{descriptive} and \textit{relational} explanations. The rubric required connecting ``observation of property changes'' to ``evidence of new substance formation,'' and subsequently to ``reasoning about chemical reaction mechanisms.'' Standard retrieval consistently fetched definitions of chemical reactions but missed the intermediate reasoning criteria that distinguish a descriptive listing of changes from a relational explanation connecting cause to effect.
Without this structural retrieval, the LLM is forced to rely on its pre-trained internal parameters to infer these missing connections. This reintroduces the very risk of hallucination that RAG was intended to solve. 
Therefore, the central technical problem this work addresses is the ``Context Fragmentation" problem: how to retrieve not just relevant isolated facts, but a connected subgraph of evidence that preserves the relational dependencies essential for validating complex student arguments.

\begin{table*}[h]
\centering
\caption{Performance Comparison on NGSS Grading Tasks (Accuracy). Best results are highlighted in \textbf{bold}.}
\label{tab:main_results}
\resizebox{0.85\textwidth}{!}{%
\begin{tabular}{@{}lcccccccccccc@{}}
\toprule
 &\multicolumn{3}{c}{\textbf{NonRAG}} &\multicolumn{3}{c}{\textbf{RAG}} & \multicolumn{3}{c}{\textbf{HippoRAG}} & \multicolumn{3}{c}{\textbf{MS GraphRAG}} \\ \cmidrule(lr){2-4} \cmidrule(lr){5-7} \cmidrule(lr){8-10} \cmidrule(lr){11-13}
\textbf{Task} & \textbf{DCI} & \textbf{SEP} & \textbf{CCC} & \textbf{DCI} & \textbf{SEP} & \textbf{CCC} & \textbf{DCI} & \textbf{SEP} & \textbf{CCC} & \textbf{DCI} & \textbf{SEP} & \textbf{CCC} \\ \midrule
Task 1 & 0.233 & 0.367 & 0.267 & 0.367 & 0.333 & 0.233 & \textbf{0.919} & \textbf{0.811} & 0.757 & 0.838 & 0.730 & \textbf{0.811} \\
Task 2 & 0.478 & 0.043 & 0.239 & 0.435 & 0.065 & 0.196 & \textbf{0.696} & \textbf{0.848} & \textbf{0.913} & 0.652 & 0.674 & 0.870 \\
Task 3 & 0.114 & 0.523 & 0.136 & 0.114 & 0.545 & 0.136 & 0.591 & \textbf{0.523} & 0.409 & \textbf{0.750} & 0.409 & \textbf{0.727} \\
Task 4 & 0.304 & 0.283 & 0.500 & 0.304 & 0.261 & 0.478 & 0.500 & \textbf{0.522} & \textbf{0.630} & \textbf{0.522} & 0.457 & 0.609 \\
Task 5 & 0.355 & 0.161 & 0.581 & 0.387 & 0.226 & 0.581 & \textbf{0.806} & \textbf{0.645} & \textbf{0.839} & 0.645 & 0.548 & 0.516 \\
Task 6 & 0.848 & 0.478 & 0.283 & 0.848 & 0.500 & 0.304 & 0.848 & \textbf{0.543} & \textbf{0.609} & \textbf{0.913} & 0.391 & 0.413 \\ \midrule
\textbf{Average} & 0.389 & 0.309 & 0.334 & 0.409 & 0.322 & 0.321 & \textbf{0.727} & \textbf{0.649} & \textbf{0.693} & 0.720 & 0.535 & 0.658 \\ \bottomrule
\end{tabular}%
}
\end{table*}

\section{Methodology}

The proposed framework addresses the structural limitations of flat retrieval in automated grading by evaluating two distinct GraphRAG paradigms. Unlike standard RAG, which relies solely on vector similarity, these approaches explicitly model educational content as a structured network. This section details the distinct graph construction and retrieval strategies employed: the community-based summarization of Microsoft GraphRAG and the neurosymbolic associative memory of HippoRAG.

\subsection{Community-Based Retrieval via Microsoft GraphRAG}
To capture high-level thematic structures and hierarchical relationships within the course material, we employ the Microsoft GraphRAG framework. This approach moves beyond simple entity extraction by utilizing a modular graph construction technique. The system first segments the reference text and uses an LLM to identify entities and relationships, constructing a base knowledge graph. Crucially, it then applies the Leiden community detection algorithm to partition the graph into hierarchical clusters of related concepts.For each detected community, the system generates a natural language summary that synthesizes the key information contained within that cluster. During the grading phase, rather than retrieving raw text chunks, the system performs a ``local search'' that navigates these pre-computed community summaries. This ensures that when a student mentions a specific concept, the LLM receives a context window enriched with the broader thematic implications and neighbor-descriptions of that concept, providing a holistic view of the topic area (DCI) required for grading.

\subsection{Neurosymbolic Retrieval with HippoRAG}
Complementing the community-based approach, we employ HippoRAG to address the specific multi-hop reasoning requirements of the SEP dimension. HippoRAG utilizes a neurosymbolic approach inspired by the hippocampal indexing theory of human memory. Unlike the community-summary approach of Microsoft GraphRAG, HippoRAG maintains a direct link between graph nodes and the original source text chunks.The construction phase involves extracting potential entities from the corpus and building a knowledge graph where edges represent co-occurrence or semantic dependency. During retrieval, the system employs the Personalized PageRank (PPR) algorithm. When a student submits a response, the key terms in the answer act as ``seed nodes'' on the graph. The PPR algorithm propagates activation probability from these seeds to their neighbors, effectively traversing the graph to identify relevant concepts via multi-hop paths. This allows the system to retrieve a subgraph of evidence that includes not only the concepts explicitly mentioned by the student but also the logically connected upstream prerequisites and downstream consequences, thereby reconstructing the full reasoning chain $A \rightarrow B \rightarrow C$ necessary for accurate assessment.

\subsection{Graph-Grounded Grading}
The final phase involves synthesizing the retrieved context (whether from community reports or associative subgraphs) into a coherent prompt for the grading LLM. The model is instructed to verify the student's response against this structured narrative. By explicitly grounding the evaluation in the retrieved relationships (edges) rather than just isolated facts (nodes), the system converts the grading task from open-ended generation into a constrained graph verification problem.

\section{Experiments}

To validate the efficacy of structural retrieval in automated grading, we conducted a comprehensive evaluation using the standard benchmark dataset reported in Chu et al. \cite{chu2025enhancing}. We compared our GraphRAG-driven approaches against the state-of-the-art flat RAG baseline across multiple scientific grading tasks.

\subsection{Experimental Setup}

\paragraph{Dataset}
We utilized the short answer grading dataset established in \cite{chu2025enhancing}, which is aligned with the Next Generation Science Standards (NGSS). The dataset consists of student responses across six distinct assessment tasks (task 1 through task 6). Each task is graded along three multidimensional learning distinct dimensions. DCI (disciplinary core ideas) is the evaluation of specific scientific content knowledge. SEP (science and engineering practices) is the assessment of the student's ability to engage in scientific reasoning and argumentation. CCC (crosscutting concepts) is the evaluation of connections between different scientific domains.
This multidimensional scoring makes the dataset particularly challenging, as it requires the grading system to verify not just factual recall (DCI) but also logical process (SEP) and synthesis (CCC).

\paragraph{Baselines and Models}
We benchmarked our proposed methods against the standard RAG implementation described by Chu et al. \cite{chu2025enhancing}, which utilizes flat vector retrieval to fetch scoring rubrics. We implement HippoRAG as our primary neurosymbolic retrieval approach that utilizes personalized PageRank to capture multi-hop dependencies. We also include Microsoft GraphRAG (local), which is an entity-centric approach that retrieves context based on the ego-graphs of key terms found in the student response.
We also performed an ablation study labeled \textit{-Background}, examining the performance when the retrieval is constrained or focused specifically on the background knowledge graph components.

\subsection{Main Results}

Table \ref{tab:main_results} presents the comparative performance of the baseline RAG system versus our GraphRAG implementations (including Microsoft GraphRAG and HippoRAG). The results shown in Table \ref{tab:main_results} demonstrate that incorporating structural graph information significantly enhances grading accuracy across all dimensions.

\paragraph{Analysis of Structural Retrieval}

Our HippoRAG implementation achieved the highest average performance across all three metrics, with a substantial improvement in the SEP dimension (0.649 vs. 0.322 for baseline RAG). This confirms our hypothesis that flat retrieval struggles with the reasoning chains required for SEP. By traversing the graph, HippoRAG successfully retrieves the intermediate logical steps that students implicitly rely on, allowing the LLM to verify their reasoning process rather than just keyword matching. Microsoft GraphRAG (local) also outperformed the baseline in DCI (0.720 vs. 0.409) and CCC (0.658 vs. 0.321), showing that entity-centric graph context is highly effective for factual verification, though it is slightly less effective than HippoRAG for capturing the complex procedural dependencies in SEP.

\subsection{Ablation Study: Impact of Background Knowledge}
We further investigated the impact of different retrieval configurations by isolating the background knowledge components. Table \ref{tab:ablation} details the performance of the \textit{-Background} variants for both graph methods.

\begin{table}[h]
\centering
\caption{Ablation Study on Background Knowledge Configuration}
\label{tab:ablation}
\resizebox{0.5\textwidth}{!}{%
\begin{tabular}{@{}lcccccc@{}}
\toprule
 & \multicolumn{3}{c}{\textbf{HippoRAG - Background}} & \multicolumn{3}{c}{\textbf{MS GraphRAG - Background}} \\ \cmidrule(lr){2-4} \cmidrule(lr){5-7}
\textbf{Task} & \textbf{DCI} & \textbf{SEP} & \textbf{CCC} & \textbf{DCI} & \textbf{SEP} & \textbf{CCC} \\ \midrule
Task 1 & 0.838 & 0.649 & 0.811 & 0.541 & 0.757 & 0.730 \\
Task 2 & 0.652 & 0.935 & 0.978 & 0.587 & 0.761 & 0.891 \\
Task 3 & 0.818 & 0.386 & 0.773 & 0.795 & 0.455 & 0.750 \\
Task 4 & 0.391 & 0.413 & 0.609 & 0.478 & 0.478 & 0.652 \\
Task 5 & 0.548 & 0.548 & 0.452 & 0.613 & 0.581 & 0.452 \\
Task 6 & 0.891 & 0.370 & 0.326 & 0.870 & 0.435 & 0.348 \\ \midrule
\textbf{Average} & 0.690 & 0.550 & 0.658 & 0.647 & 0.578 & 0.637 \\ \bottomrule
\end{tabular}%
}
\end{table}

The ablation results highlight a distinct trade-off. While the full HippoRAG system (average SEP 0.649) outperforms the background-only variant (average SEP 0.550), the MS GraphRAG (local) - Background configuration achieved a higher SEP score (0.578) than its standard counterpart (0.535). This suggests that for certain entity-centric retrieval methods, focusing the context explicitly on background knowledge can enhance the model's ability to evaluate scientific practices, potentially by reducing noise from less relevant rubric details. However, for factual consistency (DCI), the full graph integration consistently yields the best results.

\section{Discussion}
The experimental results provide strong evidence that structural retrieval methods substantially outperform flat vector-based approaches for automated short answer grading. In this section, we analyze representative case studies that illustrate the specific mechanisms by which GraphRAG improves grading accuracy, particularly in scenarios where baseline methods fail systematically.

\subsection{Case Study 1: No-RAG Failure in SEP Assessment}

The most striking performance gap between methods occurs in the Science and Engineering Practices (SEP) dimension of Task 2 (Chemical Reactions in Soap Making). The No-RAG baseline achieved only 4.3\% accuracy on this dimension, essentially random performance, while HippoRAG achieved 84.8\% accuracy, representing a 19$\times$ improvement.

\paragraph{The Assessment Challenge}
The SEP dimension requires the grader to distinguish between \textit{descriptive} explanations (merely listing observations) and \textit{relational} explanations (connecting observations to underlying scientific principles). Consider the following student response:

\begin{quote}
\vspace{-16pt}
\textit{``Your Claim: Yes a chemical reaction did occur.}\\
\textit{Evidence: Salubility in water. Before: \#1=Yes \#2=No. After: \#1=No \#2=yes.}\\
\textit{Reasoning: The salubility changed in water.''}
\vspace{-16pt}
\end{quote}

This response should receive a score of 0 (descriptive) because while the student correctly identifies a property change, they fail to \textit{explain why} this change indicates a chemical reaction occurred. The reasoning merely restates the observation without connecting it to the formation of new substances with different properties.

\paragraph{Why No-RAG Fails}
Without access to grading rubrics, the LLM must rely on its pre-trained knowledge to evaluate whether an explanation is ``relational.'' The model's general understanding of chemical reactions leads it to accept any mention of property changes as evidence of scientific reasoning. For instance, when evaluating:

\begin{quote}
\vspace{-12pt}
\textit{``Evidence: All the numbers on density, mass, and melting point are all different which means it created something new with new properties.}\\
\textit{Reasoning: If it makes something new with new properties, that means it's a chemical reaction.''}
\vspace{-12pt}
\end{quote}

The No-RAG system incorrectly classifies this as relational (score 1) because it \textit{sounds} like scientific reasoning. However, according to the rubric, this is still descriptive: the student is applying a memorized rule (``new properties = chemical reaction'') rather than constructing a relational explanation that connects specific property changes to the molecular rearrangement process.

\paragraph{Systematic Misalignment}
Our analysis revealed that the No-RAG baseline consistently exhibited a pattern of \textit{over-acceptance}: the model's prior knowledge about what constitutes ``good'' scientific explanation conflicts with the specific rubric criteria. The rubric requires students to go beyond identifying changes to explaining the \textit{mechanism} connecting those changes: a nuance that generic pre-training cannot capture. This explains why SEP accuracy (4.3\%) was dramatically lower than DCI accuracy (47.8\%) for the same task: factual content (DCI) aligns better with the model's training distribution, while procedural reasoning (SEP) requires task-specific criteria.

\subsection{Case Study 2: GraphRAG Success through Structural Context}

In contrast to the No-RAG failure, HippoRAG achieved 84.8\% accuracy on Task 2 SEP by retrieving structured grading criteria that explicitly define the distinction between descriptive and relational explanations.

\paragraph{Retrieved Context Analysis}
When processing the same student responses, HippoRAG retrieved interconnected passages that form a coherent grading framework:

\begin{enumerate}
\item \textbf{Assessment Criteria Node}: ``A Level 0 response (coded as 0) shows incomplete understanding... The student may not correctly identify the reaction as chemical, confuse physical and chemical changes, or not recognize that new substances are formed with different properties.''

\item \textbf{Level 0 Example Node}: ``Yes a chemical reaction did occur. Salubility in water Before: \#1=Yes \#2=No After: \#1=No \#2=yes. The salubility changed in water.'' This response notes a property change but doesn't explain why this indicates a chemical reaction or mention new substances.

\item \textbf{Key Concepts Bridge Node}: ``When evaluating student understanding of chemical reactions, look for: (1) Recognition that new substances are formed, (2) Identification of evidence for chemical change, (3) Understanding that the reaction cannot be easily reversed, (4) Recognition that atoms are rearranged to form new molecules.''
\end{enumerate}

\vspace{-12pt}

\paragraph{Multi-Hop Retrieval in Action}
The critical advantage of HippoRAG lies in its ability to traverse the knowledge graph to retrieve the ``Key Concepts Bridge Node'' even when the student's response does not explicitly contain the terminology in that node. The Personalized PageRank algorithm propagates activation from the seed concepts (``chemical reaction,'' ``property change'') through the graph edges to reach the bridging criteria. This multi-hop traversal is precisely what flat retrieval cannot achieve: standard vector similarity would not retrieve the ``Key Concepts'' passage because it shares limited lexical overlap with the student's response.

\paragraph{Grounding the Decision}
With this structured context, the LLM can perform a constrained verification rather than open-ended generation. The model observes that the student's response (``The salubility changed in water'') matches the Level 0 example almost verbatim, and that the response lacks the key indicators defined in the bridge node (no mention of new substances, molecular rearrangement, or irreversibility). The grading decision becomes a graph verification task: does the student's reasoning path through the conceptual space match a valid path defined in the rubric graph?

\subsection{Case Study 3: Distinguishing Relational from Descriptive Explanations}

To further illustrate the nuanced distinction that GraphRAG enables, consider two student responses that differ subtly but critically:

\paragraph{Response A (Correctly graded as 0 - Descriptive):}
\vspace{-20pt}
\begin{quote}
\vspace{-12pt}
\textit{``Claim: A chemical reaction did happen.}\\
\textit{Evidence: The lye reacted with the oil.}\\
\textit{Reasoning: Because the lye and coconut oil heated together to make a germ-killing solid.''}
\end{quote}
\vspace{-22pt}
\paragraph{Response B (Correctly graded as 1 - Relational):}
\begin{quote}
\vspace{-12pt}
\textit{``Claim: When Mixing Coconut Oil With lye and heating them up a chemical reaction does occur.}\\
\textit{Evidence: The Melting Point of Coconut Oil is 27°C, The melting Point of lye is 318°C. The Melting point of the soap he created was 48°C. The Melting Point of Glycerol is 17°C.}\\
\textit{Reasoning: The melting point of Coconut Oil and lye before they were combined was very different than after they were mixed together with heat.''}
\end{quote}
\vspace{-12pt}
Both responses identify property changes and conclude that a chemical reaction occurred. However, Response B provides \textit{specific quantitative evidence} and \textit{explicitly connects} the before/after property differences to support the claim. Response A merely describes what happened (``heated together to make a solid'') without explaining \textit{why} this transformation constitutes evidence of a chemical reaction.

\paragraph{GraphRAG's Discrimination Mechanism}
HippoRAG's retrieved context includes the Level 1 criteria: ``The student provides specific evidence about formation of new substances with different properties.'' The graph traversal also retrieves example Level 1 responses that demonstrate the expected level of specificity. By providing this contrastive context, both what constitutes Level 0 and Level 1, the LLM can accurately position Response A as descriptive (lacks specific property values) and Response B as relational (explicitly compares specific melting points before and after).
The No-RAG and standard RAG baselines lack this contrastive grounding. Without explicit examples of both categories, these systems tend to accept any response that mentions property changes as ``relational,'' leading to systematic over-scoring.

\subsection{Implications for Educational Assessment}

These case studies reveal several important implications for the design of automated grading systems:
\vspace{-12pt}
\paragraph{Task-Specific Criteria are Essential}
The dramatic failure of No-RAG (4.3\% SEP accuracy) demonstrates that LLMs cannot reliably apply task-specific grading standards from general knowledge alone. Even sophisticated models with extensive pre-training require explicit rubric context to distinguish nuanced categories like descriptive vs. relational explanations.
\vspace{-12pt}
\paragraph{Structural Retrieval Enables Multi-Hop Reasoning}
The success of HippoRAG on SEP tasks (84.8\% vs. 4.3\% for No-RAG) is directly attributable to its ability to retrieve conceptually related but lexically distinct passages. The ``bridge nodes'' that define grading criteria are often not semantically similar to student responses, making them invisible to flat vector retrieval but accessible through graph traversal.
\vspace{-12pt}
\paragraph{Contrastive Examples Improve Discrimination}
GraphRAG's retrieval of both positive and negative examples (Level 0 and Level 1 responses) provides the LLM with decision boundaries rather than isolated definitions. This contrastive grounding is particularly important for grading tasks that require fine-grained categorical distinctions.

\section{Conclusion}
This work demonstrates that graph retrieval-augmented generation (GraphRAG) significantly outperforms standard flat vector retrieval for automated short answer grading. Our evaluation of Microsoft GraphRAG and HippoRAG on an NGSS-aligned dataset reveals substantial improvements across all grading dimensions, with HippoRAG achieving the most significant improvement in SEP accuracy. This dramatic gap occurs because SEP assessment requires distinguishing nuanced categorical distinctions that demand explicit rubric criteria, which is the context that graph-based multi-hop retrieval can provide but flat vector search cannot.
Our findings suggest that for higher-order assessment tasks evaluating scientific reasoning, structural retrieval is essential for achieving alignment with human expert judgment. Future work will explore domain-specific educational knowledge graphs and personalized retrieval based on anticipated student misconceptions.

\vspace{10pt}


%
\bibliographystyle{abbrv}
\bibliography{secs/ref}  
%

\end{document}